\documentclass[journal,twoside,web]{IEEEtran}
\usepackage{cite}
\usepackage{amsmath,amssymb,amsfonts}
\usepackage{algorithmic}
\usepackage{graphicx}
\usepackage{textcomp}
\usepackage{amsmath,bm}
\usepackage{soul} 
\usepackage{multirow}
\usepackage{soul,color,colortbl}
\usepackage{caption}
\usepackage[table,xcdraw]{xcolor}
\usepackage{slashbox}
\usepackage{arydshln}

\def\BibTeX{{\rm B\kern-.05em{\sc i\kern-.025em b}\kern-.08em
    T\kern-.1667em\lower.7ex\hbox{E}\kern-.125emX}}

\begin{document}

\title{Video-Based Detection and Analysis of Errors in Robotic Surgical Training}

\author{Hanna Kossowsky Lev$^{1, 2}$, Yarden Sharon$^{1, 2}$, Alex Geftler$^{*3}$ and Ilana Nisky$^{*1, 2}$% <-this % stops a space
\thanks{*This study is supported by the Israeli Science Foundation (grant 327/20), the US-Israel Binational Science Foundation (grant 2023022), the Helmsley Charitable Trust through the ABC Robotics Initiative and SimRec, both of Ben-Gurion University of Negev, Israel. Hanna is supported by the Lachish, Ariane de Rothschild and ABC Robotics fellowships.}% <-this % stops a space
\thanks{$^{1}$Department of Biomedical Engineering, Ben-Gurion University of the Negev, Be'er Sheva, Israel}%
\thanks{$^{2}$School of Brain Sciences, Ben-Gurion University of the Negev, Be'er Sheva, Israel}%
\thanks{$^{3}$The Department of Orthopedic Surgery, Soroka Medical Center, Be'er Sheva, Israel}
\thanks{$^{*}$Alex Geftler and Ilana Nisky contributed equally.}
}

\maketitle

\begin{abstract}
Robot-assisted minimally invasive surgeries offer many advantages but require complex motor tasks that take surgeons years to master. There is currently a lack of knowledge on how surgeons acquire these robotic surgical skills.  
Toward bridging this gap, a previous study followed surgical residents learning complex surgical dry lab tasks on a surgical robot over six months. Errors are an important measure for training and skill evaluation, but unlike in virtual simulations, in dry lab training, errors are difficult to monitor automatically. Here, we analyzed errors in the ring tower transfer task, in which surgical residents moved a ring along a curved wire as quickly and accurately as possible. We developed an image-processing algorithm using color and size thresholds, optical flow and short time Fourier transforms to detect collision errors and achieved a detection accuracy of approximately 95\%. 
Using the detected errors and task completion time, we found that the residents reduced their completion time and number of errors over the six months, while the percentage of task time spent making errors remained relatively constant on average. This analysis sheds light on the learning process of the residents and can serve as a step towards providing error-related feedback to robotic surgeons.
\end{abstract}


\section{Introduction}
In robot-assisted minimally invasive surgery (RAMIS), the surgeon uses surgeon-side manipulators to control a camera arm and patient-side manipulators, on which surgical instruments are mounted. 
RAMIS has become increasingly widespread over the past decades \cite{hussain_use_2014, lanfranco_robotic_2004}, using systems such as the da Vinci surgical system (Intuitive Surgical Inc., Sunnyvale, California). RAMIS offers many benefits to both patients and surgeons, including less postoperative pain and tissue damage, shorter hospital stay \cite{hussain_use_2014, motkoski_toward_2013}, reduced tremors compared to open surgeries \cite{lanfranco_robotic_2004}, and improved ergonomics compared to laparoscopic surgeries \cite{kenngott_robotic_2008, lanfranco_robotic_2004}. 

To obtain these advantages, robotic surgeons require substantial training \cite{crawford2018evolution}. Although robotic surgical skill learning has been studied \cite{narazaki2006robotic, hung2013comparative, ahmed2015development, yang2017effectiveness, abdelaal2018play, satava2020proving, chen2019objective, chang2003robotic, sinha2023current, rota2024implementation}, there are currently many unknowns about how robotic surgeons acquire the necessary skills and a long-term study on surgical skill acquisition was needed. Therefore, in \cite{sharon2025datasetanalysislongtermskill}, surgical residents came three times throughout a shift once a month for six months. They performed several surgical training tasks; video and kinematic data were recorded to study their learning processes. All the data from was \cite{sharon2025datasetanalysislongtermskill} released to enable further analyses. In the current work, we evaluate the performance in the ring tower transfer task, in which  residents moved a ring along a curved wire as quickly as possible, while minimizing collisions with the wire. 

One challenge when assessing surgical skill is the method of evaluation. Currently, much of the evaluation is subjective, where one surgeon evaluates the performance of another and awards a score \cite{vassiliou2005global, martin1997objective, goh2012global}. These scores suffer from large variability and require a great deal of time from the evaluating surgeons.
% One challenge when assessing surgical skill is the method of evaluation. Currently, much of the evaluation is subjective, where one surgeon evaluates the performance of another and awards a score, such as Global Operative Assessment of Laparoscopic Skills \cite{vassiliou2005global}, Objective Structured Assessments of Technical Skills \cite{martin1997objective}, and Global Evaluative Assessment of Robotic Skills \cite{goh2012global}. Due to their subjective nature, these scores suffer from large variability, as well as requiring a great deal of time from the evaluating surgeons. 
RAMIS offers the possibility of overcoming these limitations by providing recorded kinematic and video data. This enables the automatic computation of metrics that are indicative of surgical skill \cite{tausch2012content, kenney2009face, liang2018motion, sharon2021rate, chien2010accuracy, chen2019objective, boal2024evaluation, brentnall2024evaluation}. Metrics that have been used include task completion time \cite{tausch2012content, liang2018motion, chien2010accuracy}, path length \cite{kenney2009face}, and movement smoothness \cite{liang2018motion}. In Sharon et al. \cite{sharon2025datasetanalysislongtermskill}, residents' kinematic metrics, such as path length and rate of orientation change, \cite{sharon2021rate} were computed for the ring tower transfer task to evaluate aspects of surgical skill.

Accuracy, or error metrics, are additional important aspects of performance in many tasks \cite{liang2018motion, chien2010accuracy}, including surgery, where errors can result in patient injury \cite{sarker2005errors, dankelman2005systems, bosma2011incidence, etchells2003patient}. The value of analyzing errors has been demonstrated in many motor control studies \cite{fitts1954information, gallea2008error, reis2009noninvasive, shadmehr2010error}. Errors have been shown to facilitate different learning processes \cite{herzfeld2014memory}, distinguishing between implicit learning driven by error in the predicted outcome of movement \cite{mazzoni2006implicit, taylor2011flexible, taylor2014explicit} and explicit learning driven by performance error \cite{mcdougle2016taking, taylor2014explicit}. 
However, many of these motor control studies focus on simple tasks, and similar error metrics are challenging to compute in the more complex RAMIS tasks \cite{chang2003robotic}, leading many works to focus on other metrics, examples of which are listed above.

Developing algorithms for detecting surgical errors can serve as a step toward reducing surgical errors \cite{shao2024think, sirajudeen2024deep, li2022runtime, xu2024sedmamba}. Furthermore, it can shed light on the learning processes of surgeons, eventually enabling the pinpointing of the different learning mechanisms that underlie skill acquisition in robotic surgical training. Moreover, detecting surgical errors can allow for designing error-related feedback for RAMIS, which has the potential to improve surgical performance \cite{kluger1996effects, 11202637}. 
Strides have been made toward detecting surgical errors. In virtual simulations, such as Intuitive Surgical's SimNow simulator, analysis of errors is now available \cite{seeliger2024skill}. 
Furthermore, several works have developed algorithms for error detection in surgical tasks \cite{shao2024think, sirajudeen2024deep, li2022runtime, xu2024sedmamba}. 
However, each of these studies has limitations, such as requiring a previously labeled dataset, and detecting errors in dry lab training remains an open challenge.

In this work, we developed an image-processing algorithm for detecting collision errors in the ring tower transfer task.
We used the video data from \cite{sharon2025datasetanalysislongtermskill}\, and wrote an algorithm to detect the collisions between the ring and the towers.
Quantifying the errors can allow for studying their progression over the six-month experiment. As there is often a trade-off between speed and accuracy \cite{chien2010accuracy, fitts1954information, reis2009noninvasive, listman2021long}, we assessed completion time together with the errors to shed light on the surgical residents' learning process of the task. 
Furthermore, this algorithm can be used to label the data for future analyses.
The main contributions of our work are:
\begin{enumerate}
    \item We discuss the residents’ learning process over six months based on completion time and error results.

    \item We developed an algorithm to detect collision errors, enabling the error analyses. The general premise of the algorithm can potentially be adapted to detect collision errors in other tasks. 

    \item We will release our codes and labeled data so they can be used for future analyses.
    
\end{enumerate}

% We characterized how the residents' performance evolved in terms of errors and completion time over the six month study. 

\section{Methods}
\subsection{Dataset}
18 surgical residents with no robotic experience from the Soroka Medical Center, Israel visited the lab 18 times -- before, during, and after a 26-hour shift approximately once a month for six months. In each visit, they used the da Vinci Research Kit (dVRK) \cite{kazanzides2014open}
to complete three surgical training tasks. They sat at the surgeon console and used the surgeon-side manipulators (SSMs) to teleoperate the patient-side manipulators (PSMs). 
The setup included two static Blackfly S cameras (FLIR Integrated Imaging Solutions Inc.) that captured the task board on the patient side. The video feed from the cameras was displayed on the stereo viewer in the surgeon console. The kinematic data, and video from the left camera, were recorded. In this work, we use the video data, which was presented to the residents at 35 Hz; however, the images were not recorded at a constant rate. 

\begin{figure*}[!htb]
\includegraphics[width=\textwidth]{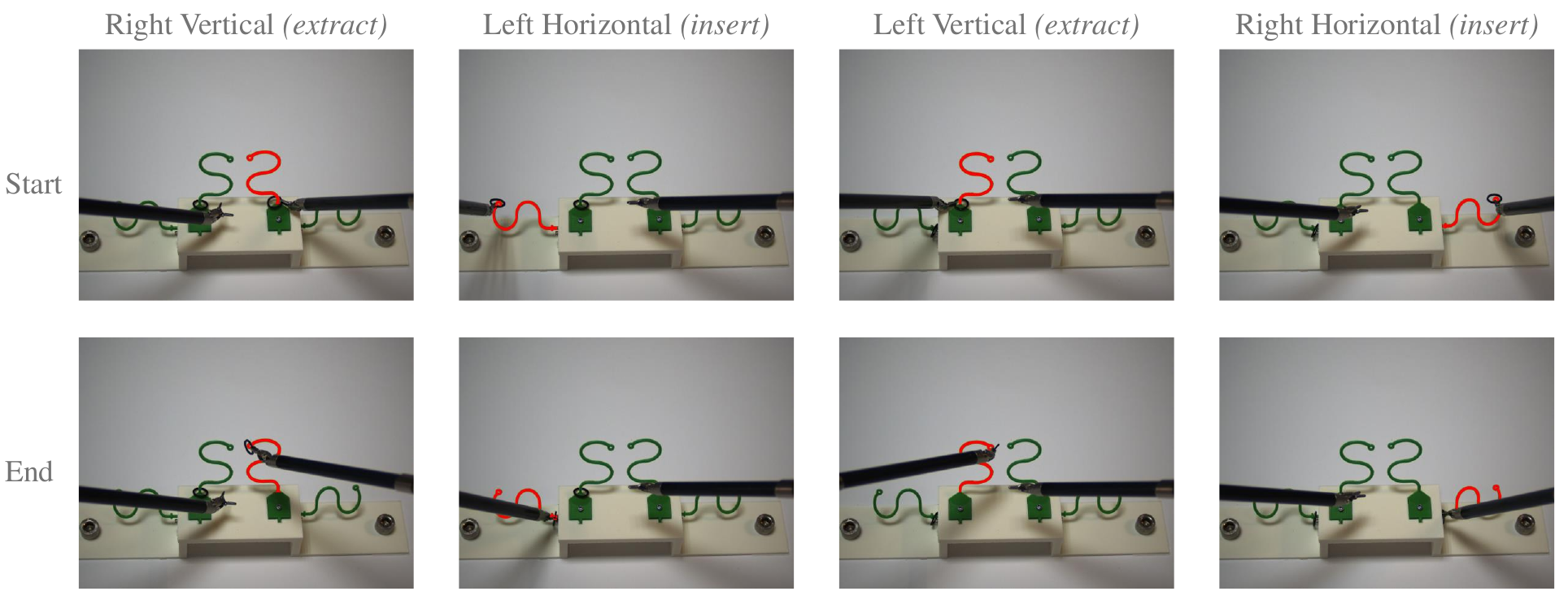}
\caption{Event Segmentation. The frames at which the residents started and ended the interactions with each of the towers were logged. For clarity, the relevant tower in each interaction is displayed here in red.}
\label{Im:segmentation}
\end{figure*}

Here, we focus on the ring tower transfer task (Fig. \ref{Im:segmentation}); the residents extracted the black ring from the right vertical (RV) tower using their right hand, transferred it to their left hand, and inserted it onto the left horizontal (LH) tower. Next, they extracted the ring from the left vertical (LV) tower with their left hand and then inserted it onto the right horizontal (RH) tower using their right hand.
The residents were asked to do so as fast as possible with minimal errors. Errors were defined as collisions between the tower and the ring. They received no feedback about their performance. Full details about the experiment and setup can be found in \cite{sharon2025datasetanalysislongtermskill}.

\subsection{Data Analysis}
\subsubsection{Event Segmentation}
18 residents each repeated the task 18 times, leading to a total of 324 repetitions. Using the video data, the frames in which the residents began and ended the interaction with each of the four towers were manually labeled. That is, extraction from the vertical towers and insertion onto the horizontal towers (Fig \ref{Im:segmentation}).

\subsubsection{Missing Data}
All 18 residents completed the task 18 times; however, one resident performed the task in the wrong order and with the wrong hand four times. Furthermore, in one visit of one resident, the task was accidentally stopped slightly early. Hence, there were five visits out of 324 that were omitted from the analyses.

\subsubsection{Error Detection}
Each collision error of the ring with one of the towers was reflected as a movement of the tower in the recorded video. Therefore, to detect each error, our main steps were to: 
(1) locate the towers in the video frames, (2) compute all movement between each two consecutive frames, (3) sum the magnitude of the velocities of all pixels located as tower pixels, and (4) use a threshold to classify the error frames.
Our goal was to detect these errors in order to analyze their progression over the 18 sessions.
We will now elaborate on each step of the algorithm. 

\begin{figure*}[!htb]
\centering
\includegraphics[width=15cm]{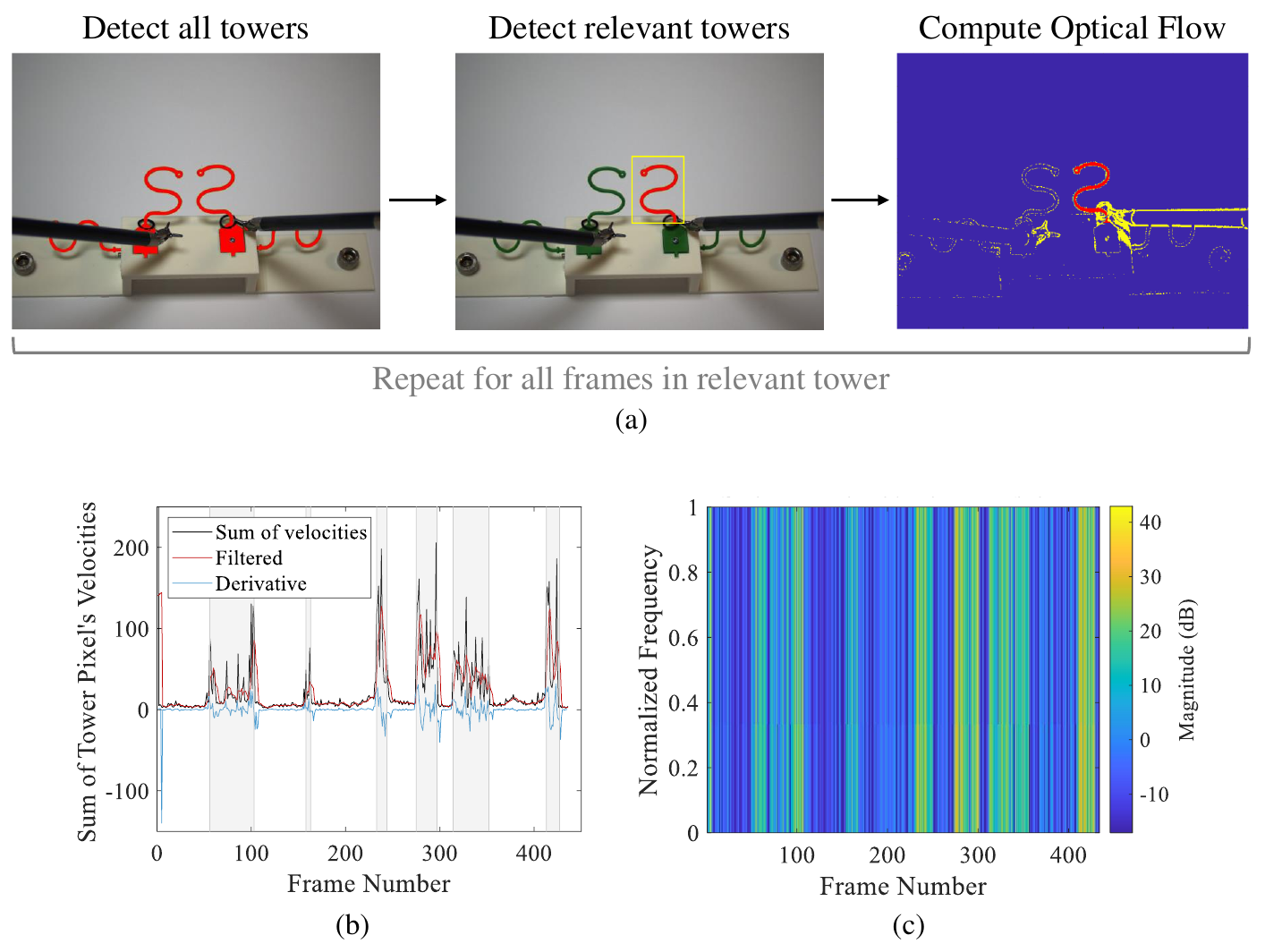}
\caption{Error Detection Algorithm. (a) Left: all pixels detected as belonging to one of the towers by the algorithm. Middle: Only the tower relevant to the current frame based on the manual event segmentation. Right: Optical flow magnitude between two consecutive frames with the detected tower pixels indicated in red. (b) The sum of the optical flow magnitude of the tower pixels (black signal), filtered (red) and derivative (blue). The shaded gray areas indicate the frames with errors. (c) Short-time Fourier transform of the derivative.}
\label{Im:thresold}
\end{figure*}

\textbf{Location of the towers:} The towers were green, and we therefore used a color threshold in the HSV space. We found all the pixels that fulfilled: (1) a hue between 70 and 130, (2) a saturation of at least 90, and (3) a value of 120 at most. Next, we imposed a minimum size threshold of 100 pixels to clean small mistakenly detected areas. Due to occlusions caused by the ring and the surgical instruments, the green towers could be detected as several separate parts. These thresholds successfully detected the towers in all the images in the residents' videos. An example can be seen in Fig. \ref{Im:thresold}(a).  
% These thresholds were set by visual examination of a small subset of images, and was found to be adequate for all the images in all the participants' videos.   

The final stage in the tower detection was to detect only the relevant tower. The thresholds detected all four green towers, however, in each frame, the resident only interacted with one of the four. To locate the relevant tower in each frame, we utilized the manual event segmentation, and the fact that the task-board was static. Therefore, although the instrument and ring moved, and could cause tower movements, each tower was limited to a neighborhood of pixels surrounding it. For each of the four towers, the algorithm returned only the detected green pixels in the neighborhood of pixels around the relevant tower based on the event segmentation (Fig. \ref{Im:thresold}(a)). This resulted in consistent detection of the relevant tower, even in cases of tower movements, occlusions, and shadows. Fig. \ref{Im:segmentation} additionally shows each of the four towers located using the algorithm by coloring them red.

\textbf{Computation of all movement between each two consecutive frames:} We used the MATLAB optical flow function with the Horn-Schunck method \cite{horn1981determining}. From this, we obtained the magnitude of each of the image pixel velocities between every two consecutive images. This signal was likely affected by the lack of a constant sampling rate; however, it was still suitable for our needs. 

\textbf{Magnitude of tower movement between every two frames: }We intersected between the two previous stages by summing the velocity magnitudes of the pixels detected as tower pixels (Fig. \ref{Im:thresold}(a)). This yielded a measurement of the total movement of the tower between every two consecutive frames in the video (Fig. \ref{Im:thresold}(b)). 
% For each interaction with a tower, we obtained the sum of the tower velocity magnitudes for each frame \ref{Im:thresold}(b).
We filtered the sum signal using a moving average filter with a window size of five frames and computed the derivative of the filtered signal.

\textbf{Threshold for detecting collision errors:} The final step was to threshold this signal. Setting the threshold in the time domain revealed several challenges. 
First, there was almost always some detected movement due to different parts of the tower being blocked by the ring and instrument in each frame, leading to different pixels being detected as tower pixels even when the tower did not move. Additionally, there were cases in which the resident was stationary for several frames during the error. Lastly, there was a large variability in the velocity values, both within, and between residents. 

Instead, we set the threshold in the frequency domain. Frames with tower movements were characterized by higher frequencies in the preprocessed signal. To obtain frequency information as a function of time, we computed short time Fourier transforms of the derivative signals (Fig. \ref{Im:thresold}(c)). This transformation is characterized by a trade-off between temporal resolution and frequency resolution. We aimed to locate the timing of the errors as precisely as possible, however we did not need information about the precise frequency content. Therefore, we chose a window size of three, leading to three frequency bands. As the signal is comprised only of real values, the magnitude of the frequency domain is symmetric. 
The magnitude in the higher frequency band needed to pass 20dB to be considered a tower movement frame.

Lastly, we set several rules to clean the thresholded signal. The first was that a detected movement in the first 10 frames was only considered a movement if it continued also in the following five frames. This was because the first frame in every video was compared to a black image. Therefore, there was commonly a detected tower movement at the beginning, which was real in some cases, and an artifact in others. Next, all detected moves within 10 samples of each other were combined. This was to cope with cases in which the tower did not significantly move between two consecutive frames during an error. Third, all lone detected movement samples in a neighborhood of five samples were removed. Fourth, if there was a detected tower movement in the last 10 samples of an interaction with one of the vertical towers, the error lasted until the end of the interaction. We did not implement this for the horizontal towers, as the final part of the interaction was to place the ring on the tower, consistently leading to an intentional tower movement. Lastly, all cases of robot crashes were noted (this occurred a total of nine times during the interaction with one of the four towers), and the frames between the beginning of the crash and the return to the task were discarded from the data in all our analyses. 

\textbf{Removal of End of Horizontal Tower Data:}
The residents placed the rings near the end of the horizontal towers. Hence, this was the only portion of the task in which a tower movement did not reflect an error, however, was still detected as one by the algorithm. Including this in the data introduces noise, and we therefore removed it as described in the following.

In an additional algorithm, we detected the black ring using color and size thresholds. Furthermore, as the ring is on the tower during the entire interaction with the tower, a minimal distance between each detected ring pixel and the nearest detected tower pixel was set. We do not elaborate on the details of this algorithm here, however, the codes and results are available in the GitHub repository. 

To identify when the residents moved to the straight part of the horizontal tower, where the ring was placed, we set a limit on the horizontal axis. This was because collisions with the horizontal towers led to tower movements primarily in the vertical axis. We set several such limits, creating separate tower segments, the last of which was the straight part at the end where residents were asked to place the ring. As collisions in this segment do not reflect errors, rather the desired placement of the ring on the tower, we removed the data from this segment from the analyses. 

\textbf{Algorithm Analysis:} Our algorithm outputted the frames with detected collisions for each resident's interaction with each of the four towers. 
To check our algorithm, we displayed the results on videos. Our code used the automatically detected tower movements and created a copy of the recorded task video while coloring the tower red every time a tower movement was detected (Video 1). 
We found that our algorithm worked well for most of the collisions. To account for occasions of missed, or wrongly detected collision errors, we wrote a code enabling us to run through the video and correct the automatic labels with a simple button press when necessary. This manual check enabled us to quantify our algorithm's ability to accurately detect the collision errors. 
To quantify its detection performance, we compared the frames detected as error frames by the algorithm with any corrections made. We computed the number of true positive (TP), true negative (TN), false positive (FP), and false negative (FN) frames for each tower and each resident.

Our algorithm achieved an accuracy of 94.79\%, a TPR (true positive rate, the ability to detect an error when there is one) of 91.98\%, a TNR (true negative rate, the ability to recognize that there is no error when there indeed is not a collision) of 98.93\%, and an F1 score of 95.46\%. These values demonstrate the high ability of the algorithm to accurately detect the collision errors. However, to allow for precise analysis of the residents' learning processes, we used the corrected versions of the labels.

All the codes and created labels are available at: https://github.com/Bio-Medical-Robotics-BGU/Error-Detection/tree/main.

\subsubsection{Surgical Learning Analysis}
Following the detection of the collision errors, we analyzed the surgical residents' performance over the six shifts and three timings relative to the shifts (before, during and after). As the residents were asked to perform the task as fast as possible while minimizing collision errors, we computed one time metric and two error metrics.

\begin{equation}
    Completion \; Time = T_{RV} + T_{LH} + T_{LV} + T_{RH},
\end{equation}
where $T$ is the time it took to complete the insertion or extraction from each tower in seconds. Each of these times was computed using the timestamps of the frames manually segmented as the start and end of the interaction with each tower. The timestamp of the start frame was subtracted from that of the end frame, and the times of the four towers were summed. Hence, the $Completion \; Time$ reflects the interaction time with the four towers, however, unlike in \cite{sharon2025datasetanalysislongtermskill} does not include the time spent transferring the rings between hands or the positioning at the start of each tower, as these data were not included in the current work.

\begin{equation}
    Number \; of \; Errors = N_{RV} + N_{LH} + N_{LV} + N_{RH},
\end{equation}
where $N$ is the number of separate collisions into each of the four towers.

\begin{equation}
    Error \; Percentage = \frac{ET_{RV} + ET_{LH} +  ET_{LV} + ET_{RH}}{Completion \; Time},
\end{equation}
where $ET$ is the total time spent colliding with each tower. The error time for each tower was computed by subtracting the first timestamp of the error from the last timestamp of the error. These times were then summed for each tower, and then for all four towers. This total error time was then divided by the $Completion \; Time$ to obtain the $Error \; Percentage$.

\subsubsection{Statistical Analyses}
We used three two-way repeated measures ANOVA models to assess the progression of each of our three metrics between months and sessions. In each model, the independent variables were shift (categorical: 1-6), timing relative to the shift (categorical: before, during, after), and resident (random). We also tested the interaction between the shift and the timing relative to the shift to assess if the effect of the timing differed between the shifts. The dependent variables were each the $Completion \; Time$, $Number \; of \; Errors$, and $Error \; Percentage$ in turn. For the $Completion \; Time$ and $Number \; of \; Errors$, we log-transformed the data to perform the analysis on normally distributed data. We used Mauchly's test to check the assumption of sphericity and used Greenhouse-Geisser to adjust the degrees of freedom in the event of violations.
The two participants with missing data were removed from the statistical analyses.
We performed the statistical analyses using the MATLAB Statistics Toolbox. Statistical significance was determined at the 0.05 threshold.

\section{Results}
% \subsection{Error Detection}
% Comparing the frames detected as errors by our algorithm with any manual corrections made allowed us to quantify the detection ability of our algorithm. 
% We achieved an accuracy of 94.79\%, a TPR (true positive rate, the ability to detect an error when there is one) of 91.98\%, a TNR (true negative rate, the ability to recognize that there is no error when there indeed is not a collision) of 98.93\%, and an F1 score of 95.46\%. These values demonstrate the high ability of the algorithm to accurately detect the collision errors. 

\begin{figure*}[!htb]
\centering
\includegraphics[width=\linewidth]{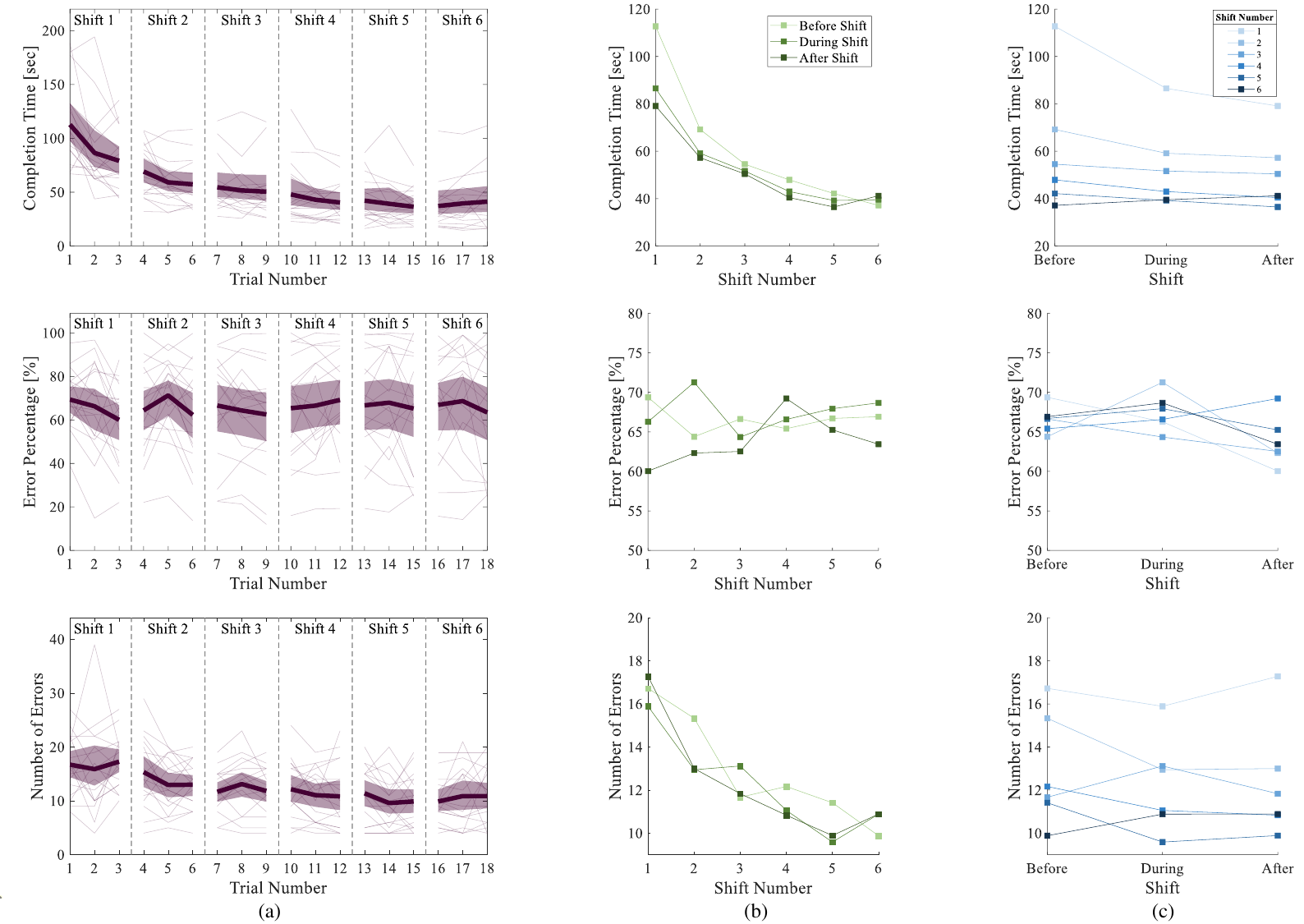}
\caption{Time and Error Results. (a) The results of all 18 residents in the 18 visits are shown in pale purple. The dark line shows the average, and the shaded area is the 95\% confidence interval, of all the residents. (b) The average results of all the residents as a function of the shift number. The three shades of green show the three timings relative to the shift (before, during and after). (c) The average results of all the residents as a function of the timing relative to the shift. The six shades of blue show the six shifts. 
The top row shows the $Completion \; Time$, the middle row is the $Error \; Percentage$ and the bottom row is the $Number \; of \; Errors$. }
\label{Im:results}
\end{figure*}

% \subsection{Surgical Learning Analysis}
Using the detected errors, we computed the completion time and error metrics for all 18 residents (Fig. \ref{Im:results}). Fig. \ref{Im:results}(a) shows the individual results of all the residents, as well as the average trend across the 18 visits. We observed a significant decrease in $Completion \; Time$, both between shifts and the timing relative to the shifts (Table \ref{tab:statistics}). This is further shown in Fig. \ref{Im:results}(b) and Fig. \ref{Im:results}(c), which are averaged across the shift, and the timing relative to the shift, respectively. This decrease reflects that, on average, the surgical residents learned to complete the task faster as the experiment progressed. The interaction between the shift and timing relative to the shift was also significant, indicating that the improvement rates in the $Completion \; Time$ between the three timings relative to the shift differed between the six shifts. 

When examining the error metrics, we observed a relatively constant average $Error \; Percentage$. There was no significant difference between shifts, reflecting that residents spent a similar percentage of time colliding with the towers in the six months. This demonstrates a lack of improvement in terms of reducing errors. There was a significant effect of the timing relative to the shift on the $Error \; Percentage$, however, the effects were small (before: $65.12 \pm 21.34\%$, during: $66.22 \pm 22.56\%$, and after: $61.30 \pm 23.03\%$). Lastly, the $Number \; of \; Errors$ significantly decreased between the six shifts, but not between the three timings relative to the shift. Taken together, the results show that the residents learned to complete the task faster, and began with making a larger number of shorter errors. With time, they made fewer separate errors, however, the portion of the time spent making the errors remained relatively constant. The statistical results are reported in Table \ref{tab:statistics}.

\begin{table}[htb!]
\caption{Statistical Time and Error Results}
    \label{tab:statistics}
\resizebox{\columnwidth}{!}{
\begin{tabular}{|cc|c|c|c|}
\hline
\multicolumn{2}{|c|}{} &
  \begin{tabular}[c]{@{}c@{}}Shift Number\\ (1/2/3/4/5/6)\end{tabular} &
  \begin{tabular}[c]{@{}c@{}}Timing Relative \\ to the Shift \\ (Before/During/After)\end{tabular} &
  Interaction \\ \hline
\multicolumn{1}{|c|}{} &
  F (df1, df2) &
  \cellcolor[HTML]{E0E0E0}54.511 (2.394, 35.914) &
  \cellcolor[HTML]{E0E0E0}5.558 (2, 30) &
  \cellcolor[HTML]{E0E0E0}2.509 (5.048, 75.723) \\ \cdashline{2-5}
\multicolumn{1}{|c|}{\multirow{-2}{*}{Completion Time}} &
  p &
  \cellcolor[HTML]{E0E0E0}\textless{}0.001 &
  \cellcolor[HTML]{E0E0E0}0.009 &
  \cellcolor[HTML]{E0E0E0}0.037 \\ \hline
\multicolumn{1}{|c|}{} &
  F (df1, df2) &
  0.102 (2.367, 35.502) &
  \cellcolor[HTML]{E0E0E0}4.381 (2, 30) &
  1.124 (10, 150) \\ \cdashline{2-5}
\multicolumn{1}{|c|}{\multirow{-2}{*}{Error Percentage}} &
  p &
  0.93 &
  \cellcolor[HTML]{E0E0E0}0.021 &
  0.348 \\ \hline
\multicolumn{1}{|c|}{} &
  F (df1, df2) &
  \cellcolor[HTML]{E0E0E0}14.049 (2.679, 40.178) &
  1.235 (2, 30) &
  1.124 (10, 150) \\ \cdashline{2-5}
\multicolumn{1}{|c|}{\multirow{-2}{*}{Number of Errors}} &
  p &
  \cellcolor[HTML]{E0E0E0}\textless{}0.001 &
  0.305 &
  0.348 \\ \hline
\end{tabular}
}
\end{table}

\begin{figure}[!htb]
\centering
\includegraphics[width=\linewidth]{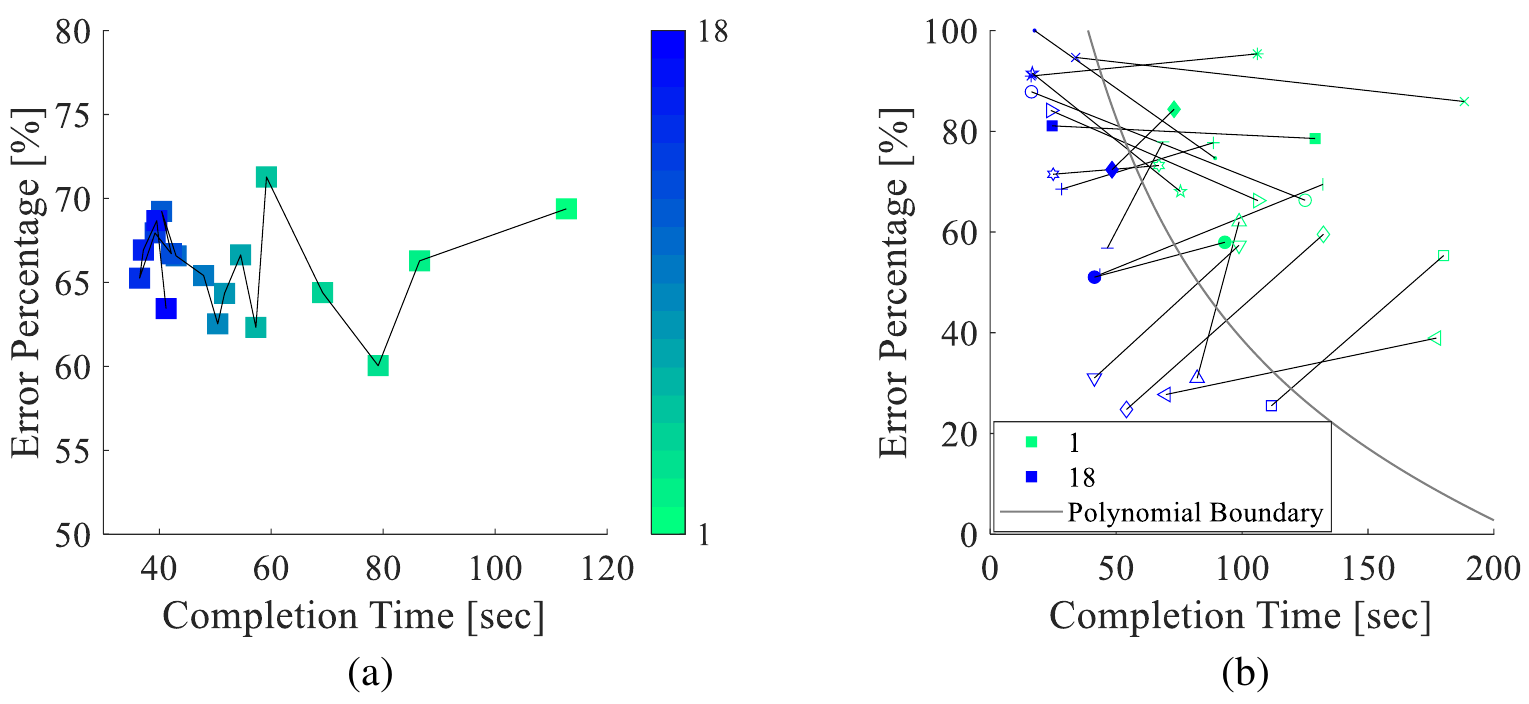}
\caption{Time-Error Analysis. (a) The average $Error \; Percentage$ as a function of the average $Completion \; Time$ across all 18 residents. The green shades indicate earlier sessions, and blue shades are the later ones, as shown in the color bar. (b) The $Error \; Percentage$ as a function of $Completion \; Time$ for each of the 18 residents in the first (green) and last (blue) visit. Each resident is shown by a different marker, and the first and last visits of each resident are connected with a line. To further demonstrate the improvement, the gray polynomial line splits the time-error space, such that all residents began on the right of the line and ended on the left. }
\label{Im:te}
\end{figure}

Fig. \ref{Im:te} further illustrates the learning process in the time-error space. In terms of error metrics, we used the $Error \; Percentage$, as it describes the amount of task time spent in error, whereas the $Number \; of \; Errors$ does not reflect their magnitude in the task. Fig. \ref{Im:te}(a) shows the average $Error \; Percentage$ of all the residents as a function of the average $Completion \; Time$ for each of the 18 visits. This illustrates the average improvement in the $Completion \; Time$ with a lack of deterioration in the $Error \; Percentage$. 

Fig. \ref{Im:te}(b) shows each resident's first and last $Error \; Percentage$ as a function of $Completion \; Time$, illustrating the individual process of each resident. We used a polynomial line to split the time-error space to demonstrate that all residents began on the right side of the line, and ended on the left. This line is similar to the Speed-Accuracy Tradeoff Functions shown in \cite{reis2009noninvasive}, characterizing the relation between the two. All the green symbols, representing the first visit, show a higher $Completion \; Time$ relative to the blue symbols (last visit), illustrating that all of the residents learned to complete the task faster over the six months. Many of the residents also showed some improvement in the $Error \; Percentage$, which taken together with the improvement in $Completion \; Time$ shows overall learning and improvement in both the time and error aspects of the task. However, there were six residents who improved their $Completion \; Time$ but increased their $Error \; Percentage$.

\section{Discussion}
We studied how surgical residents' performance in the ring tower transfer task evolved over a six month period and assessed the effect of a 26-hour shift on the learning process. To this end, we
developed an image-processing algorithm to detect the collision errors in this surgical training task. Using the results, we computed error metrics, and we additionally computed task completion time to provide a more comprehensive assessment of performance. We discovered that the residents improved their task completion time both in the sessions surrounding a single shift and across the six-month period. Furthermore, on average, this did not come at the expense of increasing their percentage of errors over time.

Many various metrics can be computed to assess different aspects of performance. In Sharon et al. \cite{sharon2025datasetanalysislongtermskill}, kinematic metrics indicative of surgical skill, such as path length and rate of orientation change \cite{sharon2021rate} were computed. Analyzing the residents’ errors in this work complements the previously evaluated kinematic metrics. We examined the residents' error metrics together with their completion time as there is often a trade-off between the two \cite{chien2010accuracy, fitts1954information, reis2009noninvasive, listman2021long}. We found that, on average, the residents spent a relatively constant percentage of time colliding with the tower over the six months, whereas they improved their task completion time. Comparing each resident's individual performance in the first and last of the 18 visits showed that 12 improved in both metrics, reflecting learning and improvement over this long-term study. However, six residents showed an increase in error percentage, raising the question of whether they improved, or rather traded off performance in one aspect of the task for another. Further indications of performance improvement were documented in \cite{sharon2025datasetanalysislongtermskill}, which reported improvements in kinematic metrics over the six-month period.

We also considered the potential effect of the 26-hour shift on the residents' learning process. We found that their completion times decreased both between shifts and between the timings relative to the shifts. That is, despite being after a 26-hour shift, the residents still improved. Studies have shown that sleep can improve motor skills \cite{walker2002practice, al2013sleep}, and it is possible that an even greater improvement would have been observed in the absence of fatigue. The reduction in completion time is consistent with the reported improvement in path length across the timings relative to the shifts, whereas the kinematic metric rate of orientation change did not reflect this improvement \cite{sharon2025datasetanalysislongtermskill}.
Furthermore, residents sometimes improved their completion time between the last visit of one month and the first visit of the following month, despite not interacting with a surgical robot during the interim.
This may indicate offline learning \cite{al2013sleep, dayan2011neuroplasticity}, but future studies are needed to determine if such offline learning indeed took place.

We additionally learned that the characteristics of the errors changed over the six months. Our observed decrease in completion time and number of errors, together with the relatively constant average error percentage, indicates that many residents began with making a larger number of shorter errors, and progressed to making fewer errors, each of which spanned a larger portion of the tower. This might reflect that at the beginning some residents were more careful, and with time, put a larger emphasis on completion time while making less of an effort to minimize collisions. Another possibility is that their increase in task speed over time, and the time needed to initiate a correction to the errors, may not have allowed for stopping the error quickly enough to decrease the error percentage \cite{briere2011automatic}.

The residents were asked to complete the task as quickly as possible while minimizing errors; however, they received no feedback on their performance. Feedback has been shown to influence various aspects of performance \cite{kluger1996effects, 11202637}, and it is therefore possible that different feedback strategies would lead to different results. Among the various potential applications of error detection in surgical tasks \cite{shao2024think, sirajudeen2024deep, li2022runtime, xu2024sedmamba}, one is the development of surgical training protocols. The design of effective training protocols involves many components, one of which is the performance feedback provided to the surgeon. As one illustration, Yang et al. \cite{yang2017effectiveness} demonstrated that presenting users with videos of surgical errors made by other novices, alongside a video of an expert correctly performing the surgical task, resulted in better learning compared to watching only the expert performance. This demonstrates the importance of understanding how surgeons acquire robotic surgical skills and highlights the potential value of error-related feedback in such learning. Future studies can test whether penalizing errors leads to their decrease and examine whether this comes at the expense of the completion time, as expected based on the speed-accuracy trade-off \cite{chien2010accuracy, fitts1954information, reis2009noninvasive, listman2021long}. It is logical that, given the lack of explicit feedback, the residents in our experiment preferred to finish faster due to their busy and demanding schedules rather than take longer and attempt to reduce their errors.
However, as errors in surgery can cause serious injuries \cite{sarker2005errors, dankelman2005systems, bosma2011incidence, etchells2003patient}, there can be cases in which prioritizing error reduction over completion time could be beneficial.
 
To reduce reliance on manual annotation in this study, we developed an algorithm for the automatic detection of collision errors in the ring tower transfer task. 
Although we manually checked the algorithm and corrected the automatic labels when necessary, this was more consistent and far less time-consuming than manually labeling the data. 
We found that our algorithm achieved accuracy and F1 scores of approximately 95\%, demonstrating its ability to automatically detect collision errors in the surgical training task. 
Several previous studies have developed algorithms for automatically detecting surgical errors \cite{shao2024think, sirajudeen2024deep, li2022runtime, xu2024sedmamba}. For example, the researchers in \cite{li2022runtime} automatically detected errors such as multiple attempts and tool out of view in suturing and needle passing tasks in the JIGSAWS dataset. They used neural networks to classify surgical gestures as correct or erroneous, achieving F1 scores ranging between 75\%-95\%. Shao et al. \cite{shao2024think} detected errors in the same tasks, achieving accuracy and F1 scores of approximately 70\% on the frame level, and presented very efficient computation, allowing for real-time analysis. While showing promising results, both \cite{li2022runtime, shao2024think} used neural networks and required the data to be labeled. Our algorithm did not rely on any prior indications of when errors occurred and can serve as a tool for labeling the data for future training of networks. The high accuracy and F1 scores show the reliability of such labeling. In addition to allowing us to study the residents' errors, we will use this labeled dataset for future analyses. 

It is important to note that although the errors themselves were not labeled, the algorithm did use manual event segmentation, which allowed us to analyze the interaction with each of the four towers separately. This segmentation was conducted as part of a previous analysis of the data and was utilized in this study. However, we suggest potential ways to overcome the lack of such segmentation to enable further automation of the algorithm. First, the entire task video could have been analyzed without separation based on the tower. Frames with tower movements would likely still be characterized by higher frequencies in the velocity-magnitude signal. We hypothesize that the algorithm would still be successful, but anticipate that the performance might be impacted by noisier signals. 
Second, it may be possible to achieve the event segmentation automatically using the PSMs' kinematic data. The task board was static, such that the beginning and end of each tower interaction would occur near fixed coordinates. Additionally, the interaction with each tower involved only one hand, while the movements made between towers, which consisted of transferring the ring between the hands, involved both hands. Lastly, the order of the four towers remained constant. Hence, the beginning of the interaction with a tower could potentially be identified by correlating the arrival of the relevant PSM to the start area of the tower with near-zero velocity in the other PSM. Similarly, the end of the interaction with a tower would involve arriving at the end area of the tower, correlated with the other PSM beginning to move. 
These suggestions would increase the computational cost and would likely affect the performance to an extent; however, they offer potential avenues for fully automating detection using our algorithm.  

This algorithm could be developed further to generalize to other training tasks. Although this would require adapting certain aspects of it, such as the threshold values, we posit that the general premise may still be applicable for detecting collision errors in various tasks. 
It may be possible to use different methods, including thresholding \cite{doignon2004detection}, YOLO \cite{ragab2024comprehensive}, and additional deep learning methods \cite{kamtam2025deep} to locate the objects of interest. The movements of these objects can then be computed at the frame level using optical flow. Setting an appropriate threshold in the frequency domain can potentially provide a robust way of differentiating between occlusions and natural movements of anatomical features, and movements of the object of interest caused by collisions. Hence, although our algorithm includes task-specific aspects and thresholds, we hypothesize that it could potentially be expanded to other tasks and environments. Future work is needed to test this hypothesis. 

\section{Conclusions}
We studied the progression of surgical residents' collision errors in the ring tower transfer task over a six-month period. To accomplish this, we
developed an algorithm for the automatic detection of the collision errors, enabling faster and more consistent annotations. We additionally computed the task completion time and found that 
surgical residents learned to perform the ring tower transfer task faster over the span of six months. Many residents also reduced their errors, reflecting learning and improvement, while for several, shorter task times came at the expense of increased errors. 
Beyond enabling us to study the characteristics of the residents' learning process, this framework provides a labeled dataset for further analysis. 

% \section*{ACKNOWLEDGMENTS}
% We thank all the residents who participated in our experiment, Yoav Farawi for completing the manual event segmentation, and Tami Matus for the endless and relentless administrative help at all hours. 
% This work is dedicated to the memory of Dr. Daniel Levi Ludmir, who participated in this study and was murdered on October 7th, 2023.

\bibliographystyle{IEEEtran}
\bibliography{PhD}

\end{document}